\documentclass{article}
\usepackage{spconf,amsmath,graphicx}

\usepackage[utf8]{inputenc} 
\usepackage[T1]{fontenc}    
\usepackage{hyperref}       
\usepackage{url}            
\usepackage{booktabs}       
\usepackage{amsfonts}       
\usepackage{nicefrac}       
\usepackage{microtype}      
\usepackage{rotating}
\usepackage{multirow}
\usepackage{amssymb}%
\usepackage{amsthm}
\usepackage{xcolor}
\usepackage{array}
\usepackage{enumerate}
\usepackage{stmaryrd}
\usepackage{ifthen}
\usepackage{dsfont}
\usepackage{enumitem}

\usepackage{lipsum}

\newcommand*{\CC}{\mathcal{CC}}
\newcommand*{\im}{\mathbf{f}}
\newcommand*{\alt}{\mathbf{a}}
\newcommand*{\mt}{\mathrm{MT}}
\newcommand*{\parent}{\mathrm{par}}

\definecolor{color1}{HTML}{D73027}
\definecolor{color2}{HTML}{FC8D59}
\definecolor{color3}{HTML}{FEE090}
\definecolor{color4}{HTML}{E0F3F8}
\definecolor{color5}{HTML}{91BFDB}
\definecolor{color6}{HTML}{4575B4}

\definecolor{rouge}{rgb}{1.0,0.0,0.0}
\definecolor{blue}{rgb}{0.0,0.0,1.0}

\newcommand*{\ie}{\textit{i.e.},}
\newcommand*{\eg}{\textit{e.g.},}

\newcommand*\FinalRevision{false}
\newcommand*{\REMM}[1]{\ifthenelse{\equal{\FinalRevision}{false}}{\fbox{\begin{minipage}{0.4\textwidth}\textsl{\textcolor{rouge}{#1}}\end{minipage}}}{#1}}
\newcommand*{\REM}[1]{\ifthenelse{\equal{\FinalRevision}{false}}{\textbf{\uline{\textcolor{rouge}{#1}}}}{}}
\newcommand*{\TODO}[1]{\ifthenelse{\equal{\FinalRevision}{false}}{\fbox{\begin{minipage}{0.4\textwidth}\textsl{\textcolor{blue}{#1}}\end{minipage}}}{#1}}
\newcommand*{\rev}[1]{\ifthenelse{\equal{\FinalRevision}{false}}{\textbf{\textcolor{blue}{#1}}}{}}
\newcolumntype{M}[1]{>{\centering}m{#1}}

\newcommand*{\Cite}[1]{~\cite{#1}}

\providecommand{\NN}{\mathbb{N}}
\providecommand{\RR}{\mathbb{R}}

\newcommand*{\dbracket}[1]{\left\llbracket#1\right\rrbracket}
\newcommand*{\set}[1]{\left\{#1\right\}}

\newtheorem{prop}{Property}

\usepackage[colorinlistoftodos, textsize=tiny, textwidth=1.5cm]{todonotes}

\graphicspath{{fig}}

\title{Component Tree Loss Function: Definition and Optimization}
\name{Benjamin Perret, and Jean Cousty\thanks{This work  was  supported by the French ANR  grant ANR-20-CE23-0019.}}
\address{LIGM, Univ Gustave Eiffel, CNRS, ESIEE Paris, F-77454 Marne-la-Vall\'ee}
\begin{document}
%
\maketitle
\begin{abstract}
In this article, we propose a method to design loss functions based on component trees which can be optimized by gradient descent algorithms and which are therefore usable in conjunction with recent machine learning approaches such as neural networks. 
We show how the altitudes associated to the nodes of such hierarchical image representations can be differentiated with respect to the image pixel values. This feature is used to design a generic loss function that can select or discard image maxima based on various attributes such as extinction values. The possibilities of the proposed method are demonstrated on simulated and real image filtering.
\end{abstract}
\section{Introduction}
\label{sec:intro}
Component-trees are hierarchical image representations that are classically used to perform connected image analysis and filtering\Cite{Salembier.TIP1998,Jones:CVIU1999}. In such methods, an image is seen as the collection of the connected components of its level sets, thus offering a representation based on elements of higher semantic level, connected components instead of pixels, to design new image analysis methods. These approaches have provided efficient solutions in many image analysis domains such as feature detection\Cite{donoser2006efficient,xu2014tree}, segmentation\Cite{Salembier.TIP1998,Jones:CVIU1999,dalla2010morphological,xu2016hierarchical,robic2019self}, or object detection and proposal\Cite{girshick2014rich,teeninga2016statistical}.

However, those methods, based on topological decompositions, do not play well with recent machine learning approaches such as neural networks as their combinatorial nature is, at first sight, not well suited to optimization strategies based on gradient descent. In this context, some authors have recently proposed  topological loss functions\Cite{clough2019explicit,hu2019topology,clough2020topological,gabrielsson2020topology} that enables to enforce topological constraints in continuous optimization frameworks using notions coming from the persistent homology theory. It has also been shown that hierarchies of segmentations can also be used in such context with the introduction of an ultrametric layer\Cite{chierchia2019ultrametric}.

In this article, we propose a novel approach to use component trees, and more specifically max-trees, within continuous optimization methods. This approach is based on the observation that, in such trees, the altitude of a node (the level of the level-set where it first appears) is directly linked to the value of some pixels of the image. Hence, we study how we can back-propagate any slight  modification of the altitude of a node of the tree to a slight modification of the initial image.
We then design a component tree loss function that enforces the presence of a prescribed number of maxima in the image based on  maxima measures. 
We study how  extinction values\Cite{vachier1995extinction},  maxima measures notably used in mathematical morphology to define hierarchical watersheds\Cite{cousty2011incremental,perret2017evaluation},  can be used to modify the behavior of the proposed loss function. Finally, the method has been implemented in Pytorch and we provide preliminary results demonstrating the use of the proposed approach on simulated and real images.

The article is organized as follows. The definition of max-trees is recalled in Sec.~\ref{sec:mt}. Then, Sec.~\ref{sec:opt} presents how max-trees can be used in gradient descent algorithms and formalizes the general optimization problem which we address. In Sec.~\ref{sec:loss}, we define a component tree loss function used for maxima selection in the max-trees and we introduce different maxima measures. The experiments are presented in Sec.~\ref{sec:exp}. Finally, Sec.~\ref{sec:end} concludes the work and gives some perspectives.

\section{Max-trees}
\label{sec:mt}
In this section, we recall the definition of  max-trees\Cite{Salembier.TIP1998,Jones:CVIU1999} which is based on the decomposition of every possible upper thresholds of an image into connected components.


In the following, the image domain is represented by a finite nonempty set $V=\{v_i\}_{i\in\dbracket{1,n}}$ of cardinality $n$. The elements of $V$ are called \emph{pixels}. 
 Given any vector $\mathbf{v}$ of $\mathbb{R}^m$ with $m\in\NN^+$, the $i$-th component of $\mathbf{v}$ is denoted~$\mathbf{v}_i$.
 An \emph{image} is represented by a vector $\im\in\RR^n$ and, for any $i\in\dbracket{1,n}$, $f_i$ is called \emph{the value of the pixel $v_i$}. Note that any image can be represented as a vector by choosing an arbitrary ordering of the pixels (\eg{} a raster scan for 2d images) and that this choice does not change the results of the proposed method.

 Let $X$ be a subset of $V$, the set of connected components of $X$ is denoted by $\CC(X)$ where connected components may be defined by any appropriate mean: \eg{} by path connectivity in a graph. In this article, all the examples involving 2d images are based on a classical 8-adjacency relation on a regular square grid of pixels.  
 Let $\im\in\RR^n$ be an image, the set of connected components of $\im$, denoted by $\CC(\im)$, is defined by $\CC(\im)=\bigcup_{\lambda\in\RR} \set{\CC([\im]_{\lambda})}$ where, for any $\lambda\in\RR$, $[\im]_{\lambda}$ is \emph{the upper level set of $\im$ of level $\lambda$}: $[\im]_{\lambda}=\set{v_i \in V \mid \im_i \geq \lambda}$.
Note that the set $\CC(\im)$ is finite and can thus be indexed by integers: $\CC(\im)=\set{C_i}_{i\in\dbracket{1, m}}$, where $m$ is the number of connected components of $\im$.
Let $C_i$ in $\CC(\im)$, the \emph{altitude of $C_i$}  is defined as the largest level $\lambda$ in $\RR$ such that $C_i$ is a connected component of the upper level set of $\im$ at level $\lambda$: \ie{} $\max \set{\lambda\in\RR \mid C_i \in \CC([\im]^{\lambda})}$. 

\begin{figure}[htb]
    \centering
    \includegraphics[width=0.40\textwidth]{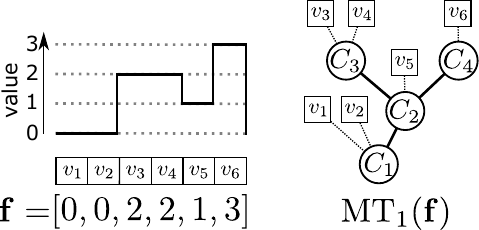}
    \caption{Max-tree example. The left figure shows a 1d image $\im\in\RR^6$ defined on the domain $v_1,\ldots, v_6$. Each of the four level sets at levels 0, 1, 2, and 3, has a single connected component $C_1,\ldots,C_4$. Those components are the nodes of hierarchy $MT_1(\mathbf{f})$ (circles) shown on the right image. The plain lines represent the parent relations between nodes.
    The proper elements of each node are depicted by squares and the dashed lines show the parent relation between those proper elements and their respective nodes, for example we have $\parent(v_4)=C_3$. The altitude vector $\alt = \mt_2(\im)$ of the max-tree of $\im$ is equal to $[0, 1, 2, 3]$, meaning for example that the altitude of the node $C_2$ is equal to 1. The two maxima of $\im$ corresponds to the leaf nodes $C_3$ and $C_4$ of the hierarchy $\mt_1(\im)$.}
    \label{fig:demoMaxTree}
\end{figure}

Let $\im\in\RR^n$ be an image. The \emph{max-tree $\mt(\im)$ of $\im$} is the pair $(\set{C_i}_{i\in\dbracket{1, m}}, \alt)$ where $\set{C_i}$ is the set of connected components of $\im$  and where $\alt$ is a vector of $\RR^m$  such that $\alt_i$  is equal to the altitude of $C_i$. The first element of the pair,  denoted by $\mt_1(\im)$, is called the \emph{hierarchy of $\mt(\im)$}. The second element of the pair, denoted by $\mt_2(\im)$, is called the \emph{altitude vector of $\mt(\im)$}. An example is given in Fig.~\ref{fig:demoMaxTree}.  
An element of the hierarchy $\mathcal{H}$ of $\mt(\im)$ is called \emph{a node of~$\mathcal{H}$}. 
A node $C_i$ of $\mathcal{H}$ is called a \emph{leaf} if there does not exists any other node included in it. There is a bijection between the leaf nodes of the hierarchy of $\mt(\im)$ and the (regional) maxima of~$\im$. 
The node $V$ includes every node of $\mathcal{H}$ and is called the \emph{root}.
For any node $C_i$ of $\mathcal{H}$, a pixel $v$ in $C_i$ that is not contained in any child of~$C_i$ is called \emph{a proper pixel of~$C_i$}. 
Any element $v$ of $V$ is a proper pixel of a unique node~$C_i$ denoted $\parent(v)$.

\section{Optimization with differentiable max-trees}
\label{sec:opt}
In this section, we first study how the altitude vector of a max-tree can be sub-differentiated, then we state the general formulation of the optimization problem which is addressed.

\textbf{Differentiable max-trees.} Trees, as combinatorial structures, are generally not suited to gradient-based optimization. However, in max-trees, the altitude of a component is  mapped to the value of some pixels of the base image: its proper pixels. Then, intuitively, a small modification of the values of those proper pixels wont change the hierarchy associated to the max-tree of the image and will produce the exact same modification of the altitude of the corresponding node of the hierarchy.
\begin{prop}
Let $\im\in\RR^n$ be an image. Let $\mathbf{\epsilon}\in\RR^n$  such that $\mt_1(\im) = \mt_1(\im + \mathbf{\epsilon})$. Then, for any node $C_i$ of $\mt_1(\im)$, the altitude of $C_i$ in $\mt(\im + \mathbf{\epsilon})$ is equal to $\alt_i + \mathbf{\epsilon}_j$ where $\alt_i$ is the altitude of $C_i$  and where $v_j$ is any proper pixel of $C_i$.
\end{prop}
This property indicates that the Jacobian of the function $\mt_2$ can be written as the matrix composed of the indicator column vectors giving the index of the node associated to any pixel of $V$ by the $\textrm{parent}$ mapping (its proper elements):
\begin{equation}
\frac{\partial \mt_2(\im)}{\partial \im} = \left[\mathds{1}_{\parent(v_1)}, \ldots, \mathds{1}_{\parent(v_n)} \right],
\end{equation}
where $\mathds{1}_{C_k}$ is the column vector of $\mathbb{R}^{m}$ equals to 1 in position~$k$, and 0 elsewhere. In a back-propagation algorithm, this means that if we have an error measure $e$ and we have already computed $\frac{\partial e}{\partial \alt}$, \ie{} how the altitude vector $\alt=\mt_2(\im)$ of the max-tree of $\im$ should be modified in order to minimize $e$,  we can then back-propagate through $\mt$ with the chain rule $\frac{\partial e}{\partial \im}=\frac{\partial \alt}{\partial \im} \frac{\partial e}{\partial \alt} $ leading to the simple formula $\left(\frac{\partial e}{\partial \im}\right)_i = \left(\frac{\partial e}{\partial \alt}\right)_{\parent(i)}$ telling how $\im$ should be modified to minimize $e$.

For example, the transpose of the Jacobian of the altitude vector $\mt_2(\im)$ of the max-tree shown in Fig.~\ref{fig:demoMaxTree} is equal to
\begin{equation*}
\bordermatrix{%
    &\im_1 & \im_2 & \im_3 & \im_4 & \im_5 & \im_6 & \cr
\alt_1 & 1 & 1 & 0 & 0 & 0 & 0 \cr
\alt_2 & 0 & 0 & 0 & 0 & 1 & 0 \cr
\alt_3 & 0 & 0 & 1 & 1 & 0 & 0 \cr
\alt_4 & 0 & 0 & 0 & 0 & 0 & 1 \cr
}.
\end{equation*}
This matrix indicates,  how the image $\im$ should be modified in order to reflect a modification of the altitude vector $\alt$ of the nodes of the max-tree of $\im$. 
For example, in order to increase the altitude $\alt_3$ of the component $C_3$ by a small value $\epsilon$, one must increase the value of $\im_3$ and $\im_4$ by this same value $\epsilon$.

\textbf{Optimization problem.} We now state a general formulation of the optimization problem that we want to solve. 
Let $\mathbf{y}\in\mathbb{R}^n$ be an image representing an observation. We are interested in solving the following optimization problem:
\begin{equation}
    \underset{\im\in\mathbb{R}^n}{\textrm{minimize}}\  J(\im;\mathbf{y}),
\end{equation}
where $J$ is a differentiable cost function involving the altitude vector $\mt_2(\im)$. As this altitude vector $\mt_2(\im)$  is differentiable with respect to the image $\im$, a local optimum of the above problem can be found by gradient descent algorithm.

\section{Maxima loss}
\label{sec:loss}
In the following, we study how to define a component tree loss imposing a topological criterion, by prescribing how many maxima should be present in the result.
The proposed approach relies on two features characterizing the maxima of an image:
\begin{itemize}
    \item \emph{a measure of saliency}: increasing this measure should reinforce the maxima and decreasing it should make it disappear; and
    \item \emph{a measure of (relative) importance}: which provides a ranking of the maxima to identify those that should be reinforced and those which should disappear.
\end{itemize}

We first introduce a generic loss function to \emph{select} a given number of maxima and to \emph{discard} the others according to these two measures. 
Then, we introduce several measures that can be used to measure the importance and the saliency of maxima.


\subsection{Ranked selection loss}
Assume that the hierarchy of $\mt(\im)$ contains $k$ maxima $\set{M_i}_{i=\dbracket{1,k}}$ (its leave nodes).  Let $\ell\in\NN^+$ be a target number of maxima. Let $\mathbf{sm}\in\RR^k$ and $\mathbf{im}\in\RR^k$ represent respectively a saliency and an importance measure on the maxima $\set{M_i}$. Then we define the ranked selection function as:
\begin{align}
    J_r(\mathbf{sm}, \mathbf{im}; \ell) & = \sum_{i=1}^{i\leq \ell} \max(m - \mathbf{sm}_{\mathbf{r}_i}, 0) + \sum_{i=\ell + 1}^{i\leq k} \mathbf{sm}_{\mathbf{r}_i} \nonumber \\
    & \textrm{ with } \mathbf{r} = \textrm{argsort}(\mathbf{im}), 
\end{align}
 where $m\in\RR$ is a constant margin, whose goal is to prevent selected maxima to grow without limit, and where $\textrm{argsort}$ is the function that associates to any vector $\mathbf{v}$ of $\RR^k$, a permutation vector $\mathbf{r}$ sorting the elements of $\mathbf{v}$ in decreasing order, \ie{} such that for any $i,j$ in $\dbracket{1,n}$, we have $i<j \Rightarrow v_{\mathbf{r}_i} \geq v_{\mathbf{r}_j}$.

\subsection{Maxima measures}
We now define maxima measures that will be used as saliency and/or importance measures in the previous loss function. Recall that the leaves of the max-tree of an image $\im$ corresponds to the maxima of this image and assume that the hierarchy of $\mt(\im)$ contains $k$ maxima $\set{M_i}_{i=\dbracket{1,k}}$; a measure on $\set{M_i}$ is then a vector of $\RR^k$.

\begin{figure}[htb]
    \centering
    \includegraphics[width=0.38\textwidth]{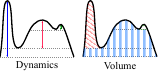}
    \caption{Illustration of the dynamics and the volume associated to the maxima of a 1d function. The dynamics of a maxima is equal to the difference of altitude between the top of the maxima and the closest level that contains another maxima of greater altitude. Similarly the volume of the maxima is equal to the surface between the top of the maxima and the closest level that contains another maxima of greater volume. With the dynamics, the most important maxima is the sharp peak on the left while, with the volume, the most important maxima is the large mount in the middle.}
    \label{fig:measures}
\end{figure}

\textbf{Highest altitude:} A simple way to measure the importance and the saliency of a maxima is to look at its highest altitude, \ie{} to the value of the pixels contained in the maxima. The highest altitude of the maxima of $\im$ is denoted $\textbf{alt}(\im)$.

\textbf{Extinction values:} Extinction values are classical maxima measure known for their robustness\Cite{vachier1995extinction}. Given a family of image filters $\set{\sigma_k}_k$  whose activity increases with $k$ (for any $k_1 \leq k_2$, we have $\sigma_{k_1}\geq\sigma_{k_2}$). The extinction value of a maxima $M_i$ of $\im$  is equal to the smallest  $k$ such that $M_i$ is not contained in any maxima of $\sigma_k(\im)$. A typical example of extinction value is the \emph{dynamics} which is based on the filtering that removes any node of the max tree that has a height (difference between the altitude of the deepest node in the subtree rooted in this node and the altitude of this node) smaller than a given threshold. The dynamics of the maxima of $\im$ will be denoted by $\mathbf{dyn}(\im)$. Another classical extinction value is the one based on the volume filter, which removes any node of the max tree whose volume is smaller than a given threshold: this measure will be denoted by $\mathbf{vol}(\im)$. In practice, computing the  extinction value of a maxima can be done efficiently in the tree by finding the saddle node associated to this maxima, that is the closest ancestor of the maxima that contains another maxima with a greater attribute value. If no such node exists (the maxima has the largest value), then we consider that its saddle node is the root of the tree.  The dynamics and volume associated to the maxima of a function are illustrated in Fig.~\ref{fig:measures}.

\begin{figure*}[htb]
    \centering
    \includegraphics[width=0.8\textwidth]{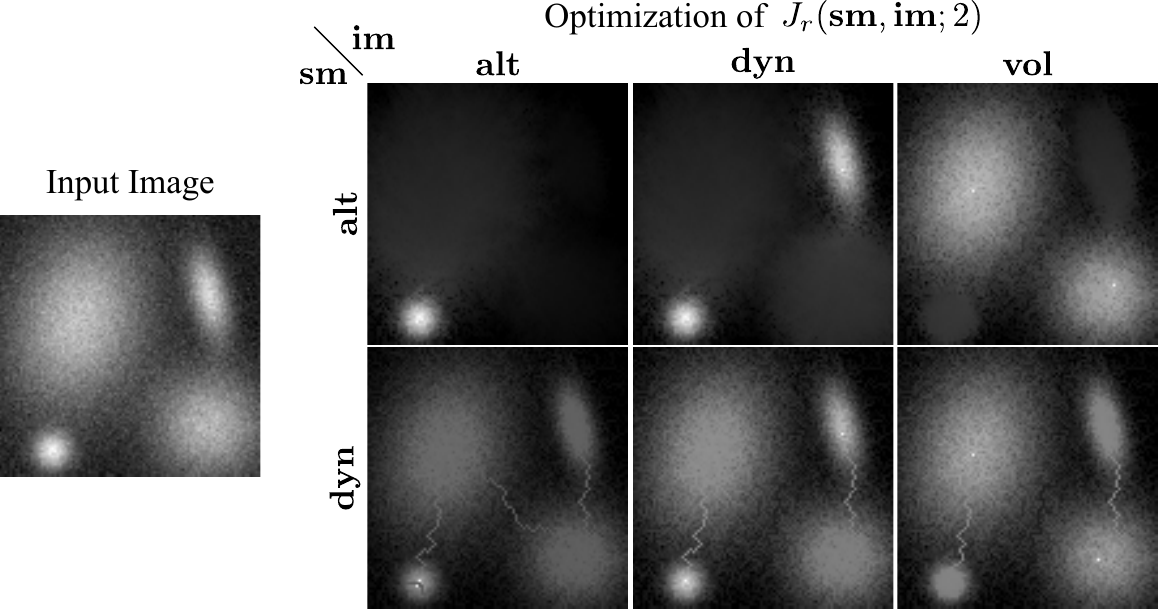}
    \caption{Optimization of the component tree loss $J_r$ on a simulated image with an objective of selecting 2 maxima. The figures on the right show the result for different combinations of maxima saliency measures ($\mathbf{sm}$) and maxima importance measures ($\mathbf{im}$).}
    \label{fig:result_demo}
\end{figure*}

\section{Experiments}
\label{sec:exp}
We demonstrate the behavior of the proposed method and the various maxima measures on a simulated image and we show how it can be combined with classical loss functions to process real images.
The method is implemented using the library Higra\Cite{perret2019higra} for hierarchical graph analysis in combination with the continuous optimization framework Pytorch\Cite{paszke2019pytorch}. In all the experiments, an Adam optimizer\Cite{AdamOptimizer} is used and the input image $\mathbf{y}$ is used as the initial solution. A Jupyter notebook containing the presented experiments is available online\footnote{\url{https://www.esiee.fr/~perretb/tmp/Component_Tree_Loss.ipynb}}.

The effect of the component tree loss  $J_r$ with the proposed importance and saliency maxima measures is demonstrated on a simulated image in Fig.~\ref{fig:result_demo}. The test image contains four maxima with different altitudes, sizes, and volumes. On the side of the importance measures, we can see that the altitude measure is not robust to noise and usually fails to select perceptually significant maxima. On the other hand, the two measures based on extinction values, the dynamics and the volume, both manage to select significant maxima: with the dynamics, the two brightest maxima are selected while with the volume, the two largest maxima are selected. Regarding saliency measures, we can see that the optimization of the saliency based on maxima altitudes leads to increasing the altitudes of the top node of the selected maxima and to raising discarded maxima. The optimization of the dynamics saliency measure is more complex as increasing/decreasing the dynamics of the maxima involves increasing/decreasing the altitude of its top node and decreasing/increasing the altitudes of its saddle node: this leads to the creation of ``bridges'' between some maxima.

Note that optimizing $J_r$ with the dynamics used both as the saliency and the importance measure of maxima is similar\Cite{boutry2019equivalence} to optimizing the barcode length of the connected components used in the ``topological loss'' based on persistent homology\Cite{clough2019explicit,hu2019topology,clough2020topological,gabrielsson2020topology}. However, as our approach does not require to compute the full persistence diagram associated to the image at each iteration of the optimization algorithm, we observe that it is dozens of times faster than a classical implementation of the topological loss\footnote{\url{https://github.com/bruel-gabrielsson/TopologyLayer}}.

\begin{figure}[htb]
    \centering
    \begin{tabular}{cc}
        \includegraphics[width=0.22\textwidth]{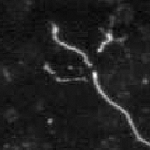} & \includegraphics[width=0.22\textwidth]{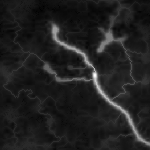} \\
         Image & Result 
    \end{tabular}
    \caption{Reconnection of a neurite image with a combination of the proposed loss $J_r$ to enforce a single maximum, a L2 data attachment term and a TV2 regularization term.}
    \label{fig:neurite}
\end{figure}

Finally, in Fig.~\ref{fig:neurite}, we show how the proposed loss function can be combined with classical loss functions used in image analysis: here we optimize the term $||\mathbf{f} - \mathbf{y}||_2^2 +  \lambda_1 J_{r}(\mathbf{dyn}(\mathbf{f}), \mathbf{dyn}(\mathbf{f}), 1) + \lambda_2 ||\nabla \mathbf{f}||_2^2$ which combines our loss based on the max-tree to enforce the presence of a single maxima with a L2 data attachment term and a total variation regularization term. We can see that we are able to successfully reconnect the different branches of the neurite.

\section{Conclusion}
\label{sec:end}
We have proposed a continuous optimization framework based on the hierarchical image representation called the max-tree. We showed how it can be used to design a component tree loss, \ie{} a regularization term, enabling to select or discard maxima in an image based on various measures. This approach can be generalized immediately to other hierarchical representations such as the min-tree or the  tree-of-shapes\Cite{ballester2003tree,geraud2013quasi}. In future works, we plan to explore more general component tree loss functions based on such hierarchical representations and their use in supervised learning methods involving deep networks.

\bibliographystyle{IEEEbib}
\bibliography{biblio}

\begin{thebibliography}{10}

\bibitem{Salembier.TIP1998}
P.~Salembier, A.~Oliveras, and L.~Garrido,
\newblock ``Anti-extensive connected operators for image and sequence
  processing,''
\newblock {\em IEEE TIP}, vol. 7, no. 4, pp. 555--570, 1998.

\bibitem{Jones:CVIU1999}
R.~Jones,
\newblock ``Connected filtering and segmentation using component trees,''
\newblock {\em CVIU}, vol. 75, no. 3, pp. 215--228, 1999.

\bibitem{donoser2006efficient}
M.~Donoser and H.~Bischof,
\newblock ``Efficient maximally stable extremal region (mser) tracking,''
\newblock in {\em IEEE CVPR}, 2006, vol.~1, pp. 553--560.

\bibitem{xu2014tree}
Y.~Xu, P.~Monasse, Th. G{\'e}raud, and L.~Najman,
\newblock ``Tree-based morse regions: A topological approach to local feature
  detection,''
\newblock {\em IEEE TIP}, vol. 23, no. 12, pp. 5612--5625, 2014.

\bibitem{dalla2010morphological}
M.~Dalla~Mura, J.A. Benediktsson, B.~Waske, and L.~Bruzzone,
\newblock ``Morphological attribute profiles for the analysis of very high
  resolution images,''
\newblock {\em IEEE TGRS}, vol. 48, no. 10, pp. 3747--3762, 2010.

\bibitem{xu2016hierarchical}
Y.~Xu, E.~Carlinet, Th. G{\'e}raud, and L.~Najman,
\newblock ``Hierarchical segmentation using tree-based shape spaces,''
\newblock {\em IEEE TPAMI}, vol. 39, no. 3, pp. 457--469, 2016.

\bibitem{robic2019self}
J.~Robic, B.~Perret, A.~Nkengne, M.~Couprie, and H.~Talbot,
\newblock ``Self-dual pattern spectra for characterising the dermal-epidermal
  junction in 3d reflectance confocal microscopy imaging,''
\newblock in {\em ISMM}. Springer, 2019, pp. 508--519.

\bibitem{girshick2014rich}
R.~Girshick, J.~Donahue, T.~Darrell, and J.~Malik,
\newblock ``Rich feature hierarchies for accurate object detection and semantic
  segmentation,''
\newblock in {\em IEEE CVPR}, 2014, pp. 580--587.

\bibitem{teeninga2016statistical}
P.~Teeninga, U.~Moschini, S.C. Trager, and M.H.F. Wilkinson,
\newblock ``Statistical attribute filtering to detect faint extended
  astronomical sources,''
\newblock {\em Mathematical Morphology-Theory and Applications}, vol. 1, 2016.

\bibitem{clough2019explicit}
J.R. Clough, I.~Oksuz, N.~Byrne, J.A. Schnabel, and A.P. King,
\newblock ``Explicit topological priors for deep-learning based image
  segmentation using persistent homology,''
\newblock in {\em IPMI}. Springer, 2019, pp. 16--28.

\bibitem{hu2019topology}
X.~Hu, F.~Li, D.~Samaras, and C.~Chen,
\newblock ``Topology-preserving deep image segmentation,''
\newblock in {\em NeurIPS}, 2019, pp. 5657--5668.

\bibitem{clough2020topological}
J.~Clough, N.~Byrne, I.~Oksuz, V.A. Zimmer, J.A. Schnabel, and A.~King,
\newblock ``A topological loss function for deep-learning based image
  segmentation using persistent homology,''
\newblock {\em IEEE TPAMI}, 2020.

\bibitem{gabrielsson2020topology}
R.B. Gabrielsson, B.J. Nelson, A.~Dwaraknath, and P.~Skraba,
\newblock ``A topology layer for machine learning,''
\newblock in {\em AISTATS}. PMLR, 2020, pp. 1553--1563.

\bibitem{chierchia2019ultrametric}
G.~Chierchia and B.~Perret,
\newblock ``Ultrametric fitting by gradient descent,''
\newblock in {\em NeurIPS}, 2019, pp. 3181--3192.

\bibitem{vachier1995extinction}
C.~Vachier and F.~Meyer,
\newblock ``Extinction value: a new measurement of persistence,''
\newblock in {\em IEEE Workshop on nonlinear signal and image processing},
  1995, vol.~1, pp. 254--257.

\bibitem{cousty2011incremental}
J.~Cousty and L.~Najman,
\newblock ``Incremental algorithm for hierarchical minimum spanning forests and
  saliency of watershed cuts,''
\newblock in {\em ISMM}. Springer, 2011, pp. 272--283.

\bibitem{perret2017evaluation}
B.~Perret, J.~Cousty, S.J.F. Guimaraes, and D.S. Maia,
\newblock ``Evaluation of hierarchical watersheds,''
\newblock {\em IEEE TIP}, vol. 27, no. 4, pp. 1676--1688, 2017.

\bibitem{perret2019higra}
B.~Perret, G.~Chierchia, J.~Cousty, S.J.F. Guimar{\~a}es, Y.~Kenmochi, and
  L.~Najman,
\newblock ``Higra: Hierarchical graph analysis,''
\newblock {\em SoftwareX}, vol. 10, pp. 100335, 2019.

\bibitem{paszke2019pytorch}
A.~Paszke, S.~Gross, F.~Massa, A.~Lerer, J.~Bradbury, G.~Chanan, T.~Killeen,
  Z.~Lin, N.~Gimelshein, L.~Antiga, et~al.,
\newblock ``Pytorch: An imperative style, high-performance deep learning
  library,''
\newblock in {\em NeurIPS}, 2019, pp. 8026--8037.

\bibitem{AdamOptimizer}
D.P. Kingma and J.~Ba,
\newblock ``Adam: {A} method for stochastic optimization,''
\newblock in {\em {ICLR}}, Y.~Bengio and Y.~LeCun, Eds., 2015.

\bibitem{boutry2019equivalence}
N.~Boutry, Th. G{\'e}raud, and L.~Najman,
\newblock ``An equivalence relation between morphological dynamics and
  persistent homology in 1d,''
\newblock in {\em ISMM}. Springer, 2019, pp. 57--68.

\bibitem{ballester2003tree}
C.~Ballester, V.~Caselles, and P.~Monasse,
\newblock ``The tree of shapes of an image,''
\newblock {\em ESAIM: Control, Optimisation and Calculus of Variations}, vol.
  9, pp. 1--18, 2003.

\bibitem{geraud2013quasi}
Th. G{\'e}raud, E.~Carlinet, S.~Crozet, and L.~Najman,
\newblock ``A quasi-linear algorithm to compute the tree of shapes of nd
  images,''
\newblock in {\em ISMM}. Springer, 2013, pp. 98--110.

\end{thebibliography}

\vfill

{
\tiny
© 2021 IEEE. Personal use of this material is permitted. Permission from IEEE must be obtained for all other uses, in any current or future media, including reprinting/republishing this material for advertising or promotional purposes, creating new collective works, for resale or redistribution to servers or lists, or reuse of any copyrighted component of this work in other works.}

\end{document}